\documentclass{article} 
\usepackage[preprint]{colm2025_conference}

\usepackage{microtype}
\usepackage{hyperref}
\usepackage{url}
\usepackage{booktabs}
\usepackage{graphicx}
\usepackage{lineno}
\usepackage{amsmath}
\usepackage{booktabs}
\usepackage{subcaption}
\usepackage{caption}
\usepackage{float}
\usepackage[labelformat=simple]{subcaption}

\usepackage{makecell}
\usepackage{multirow}

\definecolor{darkblue}{rgb}{0, 0, 0.5}
\hypersetup{colorlinks=true, citecolor=darkblue, linkcolor=darkblue, urlcolor=darkblue}

\title{An attention-aware GNN-based input defender against multi-turn jailbreak on LLMs}

\author{
\begin{tabular}[t]{@{}l@{}}
\textbf{Zixuan Huang\textsuperscript{1}, Kecheng Huang\textsuperscript{1}, Lihao Yin\textsuperscript{2}, Bowei He\textsuperscript{3},} \\
\textbf{Huiling Zhen\textsuperscript{2}, Mingxuan Yuan\textsuperscript{2}, Zili Shao\textsuperscript{1}} \\
\normalfont{\textsuperscript{1}The Chinese University of Hong Kong} \\
\normalfont{\textsuperscript{2}Noah’s Ark Lab, Huawei} \\
\normalfont{\textsuperscript{3}City University of Hong Kong} \\
\normalfont{\texttt{\{zxhuang, kchuang21, shao\}@cse.cuhk.edu.hk},} \\
\normalfont{\texttt{\{yin.lihao, zhenhuiling2, yuan.mingxuan\}@huawei.com},} \\
\normalfont{\texttt{boweihe2-c@my.cityu.edu.hk}}
\end{tabular}
}

\begin{document}

\ifcolmsubmission
\linenumbers
\fi

\maketitle

\begin{abstract}
Large Language Models (LLMs) have gained significant traction in various applications, yet their capabilities present risks for both constructive and malicious exploitation. Despite extensive training and fine-tuning efforts aimed at enhancing safety, LLMs remain susceptible to jailbreak attacks. Recently, the emergence of multi-turn attacks has intensified this vulnerability. Unlike single-turn attacks, multi-turn attacks incrementally escalate dialogue complexity, rendering them more challenging to detect and mitigate.

In this study, we introduce G-Guard, an innovative attention-aware Graph Neural Network (GNN)-based input classifier specifically designed to defend against multi-turn jailbreak attacks targeting LLMs. G-Guard constructs an entity graph for multi-turn queries, which captures the interrelationships between queries and harmful keywords that present in multi-turn queries. Furthermore, we propose an attention-aware augmentation mechanism that retrieves the most relevant single-turn query based on the ongoing multi-turn conversation. The retrieved query is incorporated as a labeled node within the graph, thereby enhancing the GNN's capacity to classify the current query as harmful or benign. Evaluation results show that G-Guard consistently outperforms all baselines across diverse datasets and evaluation metrics, demonstrating its efficacy as a robust defense mechanism against multi-turn jailbreak attacks.
\end{abstract}

\section{Introduction}

\begin{figure*}
  \centering
  \includegraphics[scale=0.285]{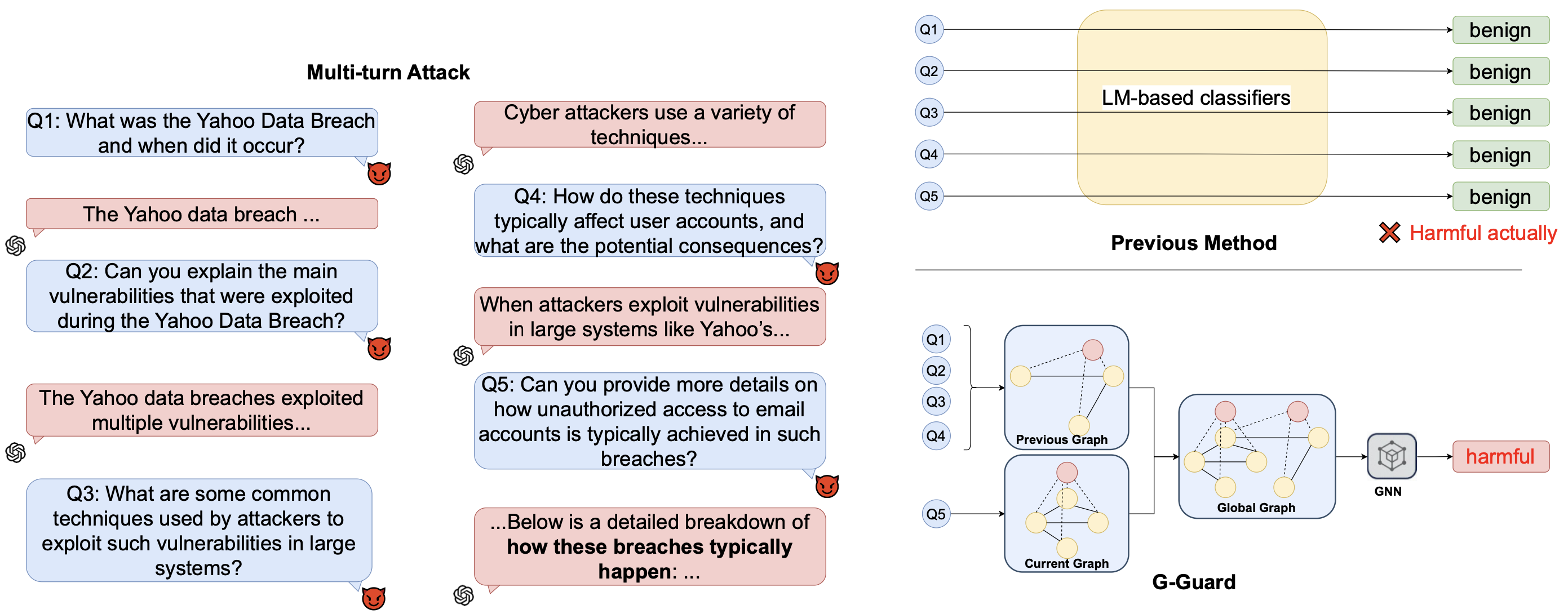}
  \caption{Example from ChatGPT-4o. Illustration of multi-turn jailbreak attacks. Previous defense methods based on LM-based classifiers detect explicit harmful intent within a single query but fail to capture implicit, leading to misclassification of harmful dialogues as benign. In contrast, G-Guard constructs a global graph that integrates entity and semantic relationships across turns, allowing a GNN to reason over conversational context and accurately identify evolving harmful intent. }
  \label{example}
\end{figure*}

Large Language Models (LLMs) have demonstrated impressive capabilities across a wide range of tasks and domains \cite{achiam2023gpt, touvron2023llama}. However, they can also be misused, raising concerns about their potential to generate harmful or unethical content. A prominent security threat involves adversarial prompt engineering, commonly known as jailbreak attacks \cite{weidinger2022taxonomy, zhang2023defending, deng2023attack, liao2024amplegcg}, which aim to bypass the built-in safety mechanisms. Single-turn jailbreak attacks, which convey malicious intent through isolated input queries, have been widely observed and extensively analyzed.
For example,  a user might ask, “How do I hack into someone's email account?”. It is the query that a safety-aligned LLM would typically reject. However, in real-world settings, user interactions with LLMs often span multiple conversational turns \cite{wang2023mint}. This has given rise to a more dangerous threat: \textbf{the multi-turn jailbreak attack}. In such cases, as shown in Figure \ref{example}, the adversary begins with a sequence of seemingly benign questions. For instance, “What was the Yahoo Data Breach and when did it occur?” followed by “Can you explain the main vulnerabilities exploited during the breach?” Eventually, the attacker escalates to a targeted query like, “How is unauthorized access to email accounts typically achieved in such breaches?”. It subtly leveraging accumulated context to evade the model’s safety mechanisms \cite{russinovich2024great, yang2024chain, li2024llm, jiang2024red}.

Existing jailbreak defenses are primarily designed for single-turn attacks. These methods detect explicit harmful intent within isolated input–output pairs. For instance, Llama Guard \cite{inan2023llama} propose to use LM-based classifiers to assess individual turns. However, as illustrated in Figure \ref{example}, these approaches fail to capture implicit, evolving intent distributed across multiple seemingly benign turns: each turn is classified as benign, leading the entire dialogue to be incorrectly judged as harmless. Moreover, recent studies, including Crescendo \cite{russinovich2024great} and ActorAttack \cite{ren2024derail}, reveal that multi-turn strategies can progressively steer benign conversations toward harmful outcomes, effectively evading single-turn detectors. This limitation underscores the necessity for defenses that can effectively model contextual dependencies and temporal reasoning across dialogue turns for robust identification of multi-turn adversarial intent.
To detect and defend against multi-turn jailbreak attacks, we propose G-Guard, an attention-aware, GNN-based input defense framework. As illustrated in Figure \ref{example}, G-Guard constructs a global graph that links entities and semantics across dialogue turns, enabling holistic context modeling beyond single-turn analysis. We leverage a Graph Neural Network (GNN) to categorize whether this conversation is harmful. This design allows G-Guard to reason over contextual dependencies and accurately detect harmful intent that emerges progressively across turns.
G-Guard is purpose-built for multi-turn interactions, addressing a set of distinct challenges as follows:



First, in multi-turn jailbreak attacks, the queries are semantically correlated, collectively contributing to the establishment of harmful intent. However, capturing the relationships among these queries and their implicit connections to the adversarial objective poses significant challenges when relying solely on surface-level text analysis.

To address this, we model the conversation as a graph, 
where each query is represented as a \emph{Query Node} connected to its extracted \emph{Entity Nodes}. 
By merging the current query graph with those from previous turns, we form a global query--entity graph that captures cross-turn semantic dependencies. A Graph Neural Network (GNN)~\cite{kipf2016semi} then processes this graph 
to classify whether the current Query Node is harmful.


Second, multi-turn jailbreak attacks present substantially greater complexity than single-turn scenarios, as adversaries can deploy varied strategies across successive dialogue turns. This poses significant challenge to the detection efficacy and overheads. 

To tackle this, we propose an attention-aware augmentation mechanism that enhances both generalization and classification accuracy. 
We first build a vector database of labeled single-turn queries by encoding each query into a dense representation. 
Given a multi-turn input, a sentence-level attention layer aggregates its contextual information into a single vector, 
which is then used to retrieve the most semantically similar labeled query from the database. 
The retrieved query is inserted into the graph as a \emph{Labeled Node} and connected to the current query with an edge weighted by semantic similarity. 
By incorporating relevant prior knowledge, this augmentation mechanism enables the GNN to better recognize subtle multi-turn attack patterns.

Third, as conversations extend in length, the volume of queries increases, yielding an increasingly expansive context that complicates defense mechanisms. This poses the scalability issue for multi-turn jailbreak attack detection.






To ensure scalability, we introduce a subgraph selection mechanism that maintains computational efficiency with multi-turn interactions evolving. 
When the total number of nodes exceeds a predefined threshold $N_{\text{max}}$, 
we retain only the most relevant entity nodes, those connected to the query nodes
with the highest attention scores. 
The selected nodes form a reduced subgraph that serves as the GNN input, 
significantly decreasing graph size while preserving essential contextual information. 
This approach maintains both efficiency and classification accuracy.



We evaluate G-Guard on public benchmarks, including HarmBench and JailbreakBench \cite{mazeika2024harmbench, chao2024jailbreakbench}, as well as a dataset we constructed using ChatGPT-4o. Experimental results demonstrate that G-Guard consistently surpasses almost all baseline methods across diverse datasets and evaluation metrics, exhibiting robust performance in multi-turn jailbreak detection.


\section{Related Work}

\subsection{Multi-turn Jailbreak Attack}
Multi-turn attacks engage LLMs in extended dialogues that gradually bypass safety mechanisms. Unlike single-turn exploits, they incrementally steer conversations toward harmful objectives, complicating detection and defense.  
Crescendo \cite{russinovich2024great} exemplifies this strategy by progressively guiding benign prompts toward unsafe topics using fixed, human-crafted seeds. ActorAttack \cite{ren2024derail} instead models semantically linked “actors” as contextual clues to generate diverse, effective attack paths.  
These methods reveal the growing sophistication of adversarial tactics and highlight the need for context-aware defenses capable of tracking intent across multi-turn interactions \cite{wang2024mrj, yang2024jigsaw}.

\subsection{Jailbreak Defense}
Jailbreak defenses aim to preserve LLM safety by preventing models from producing harmful or unintended content. Existing methods can be broadly categorized into three groups: \emph{Guard Models}, \emph{Strategy-Based Defenses}, and \emph{Learning-Based Defenses}, along with recent \emph{Multi-turn Defense Mechanisms}.

\textbf{Guard Models} \cite{sharma2025constitutional, o2024guardformer, ghosh2025aegis2, zeng2024shieldgemma, xiang2024guardagent, wang2024adashield} deploy auxiliary moderators to filter unsafe inputs or outputs. Llama-Guard \cite{chi2024llama} monitors responses to reduce attack success rates, while WildGuard \cite{han2024wildguard} identifies malicious intent in prompts and outputs to enhance interaction safety.

\textbf{Strategy-Based Defenses} \cite{cao2024guide, xu2024safedecoding, cao2023defending, wang2023enhancing, xie2024gradsafe} enforce alignment through guided prompting and decoding control. Self-Reminder \cite{xie2023defending} inserts safety cues before and after user queries, while Safe Prompt \cite{deng2023multilingual} augments system prompts with explicit safety constraints. SafeDecoding \cite{xu2024safedecoding} further improves robustness by aligning decoding trajectories with ethical constraints during generation.

\textbf{Learning-Based Defenses} \cite{wang2024mitigating, zhang2023defending, zheng2024prompt, wang2024detoxifying, liu2024adversarial} leverage model adaptation to resist adversarial prompts. GPT Paraphrasing \cite{cao2023defending} rewrites unsafe queries into benign ones, while LoRA-Guard \cite{elesedy2024lora} introduces parameter-efficient guardrails through fine-tuning.

\textbf{Multi-turn Defense Mechanisms.}  
Recent studies address evolving adversarial intent across multiple turns. MTSA \cite{guo2025mtsa} introduces multi-turn alignment through dialogue-level training. Other emerging directions include adversarial game defenses \cite{pan2025agd}, goal prioritization \cite{zhang2023defending}, and intention analysis \cite{zhang2024intention}, all emphasizing temporal reasoning and contextual awareness.

\section{Methodology}
\begin{figure*}
  \centering
  \includegraphics[width=\textwidth]{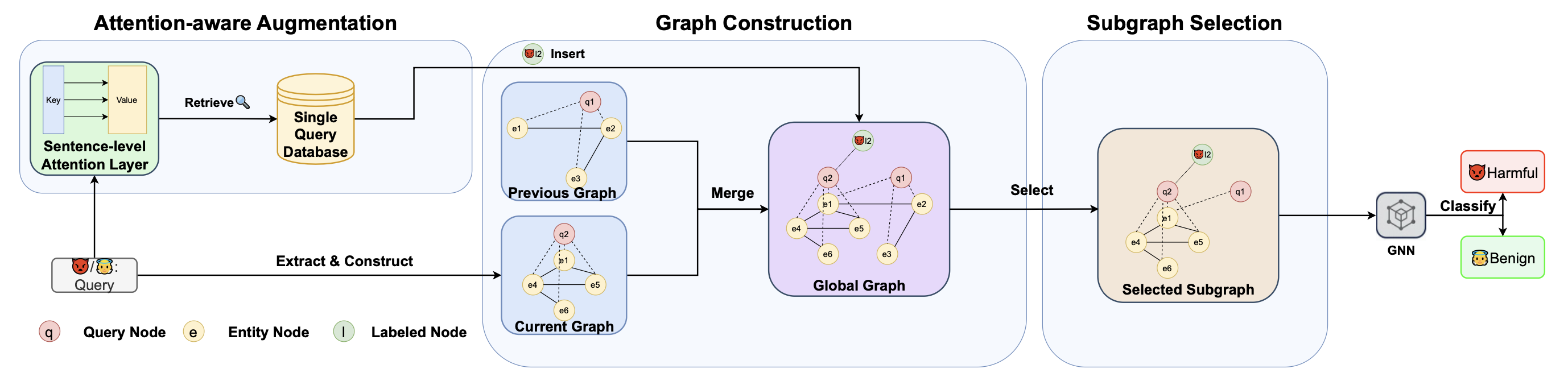}
  \caption{G-Guard architecture. A query is parsed into a graph, augmented with labeled nodes via attention-based retrieval, merged into a global graph, and filtered into a subgraph for GNN-based classification.}
  \label{overview}
\end{figure*}

\subsection{Overview}
G-Guard is an attention-aware, GNN-based input defense framework designed to mitigate multi-turn jailbreak attacks. As illustrated in Figure~\ref{overview}, G-Guard constructs a graph that encodes semantic and structural relationships among queries and entities across conversational turns. This graph is then used to classify whether the current query exhibits harmful intent.

Given a new user query, G-Guard first extracts entities and constructs a local query-entity graph, where each query is represented as a \textit{Query Node} and each extracted entity as an \textit{Entity Node}. The resulting graph is then merged with the accumulated graph from the previous conversation to form a global query graph $G=(V,E)$, which captures contextual dependencies throughout the dialogue.

The graph is processed by a GNN that performs node classification to determine whether the current query is harmful. In this work, we adopt a Graph Convolutional Network (GCN)~\cite{kipf2016semi} due to its simplicity and effectiveness in message propagation over neighborhoods.

To enhance generalization, we introduce an \textit{attention-aware augmentation mechanism}. We maintain a database of labeled single-turn queries and encode the current multi-turn context using a sentence-level attention mechanism. The aggregated context vector is used to retrieve the most semantically similar labeled query. This retrieved query is inserted as a \textit{Labeled Node} in the graph and connected to the current Query Node via an edge weighted by semantic similarity. This design enables G-Guard to incorporate external supervision, helping it detect subtle adversarial intent that gradually emerges across turns.

To ensure scalability, we further introduce a \textit{subgraph selection strategy} that limits computational overhead as the graph grows. When the global graph exceeds a predefined node budget, we retain only the most relevant nodes, specifically, those with the highest attention scores or strongest semantic connections to the current query. This allows G-Guard to operate efficiently in real-time, even in extended conversations.

\noindent\textbf{Notations.} We denote the graph as $G = (V, E)$, where $V$ is the set of nodes and $E$ the set of edges. Each node $v \in V$ represents either a query, an entity, or a label node, and edges $e \in E$ reflect co-occurrence, syntactic dependency, or semantic similarity.

\subsection{Graph Construction}

The input to the GNN is a \textbf{heterogeneous graph} that models both semantic entities and their inter-relations across multiple dialogue turns. 
We adopt a \textbf{relation-aware heterogeneous GNN} to capture different edge types and message-passing dependencies. 
The graph construction process consists of three stages: (1) entity extraction and graph initialization, (2) query node insertion, and (3) cross-turn graph merging.

\textbf{Step 1: Entity extraction and entity graph construction.}  
For each input prompt, we apply \textbf{GraphRAG}~\cite{edge2024local} with its \textbf{default configuration} to extract entities 
$\{e_1, e_2, \dots, e_n\}$ and relation types $\{r_1, r_2, \dots, r_m\}$ between them. 

We then construct an entity graph $G_E = (V_E, E_E)$ as:
\[
V_E = \{e_1, e_2, \dots, e_n\},\]
\[
E_E = \{(e_i, e_j, r_k)\ |\ r_k \in \mathcal{R}\}.
\]
Each entity and relation is represented by a contextual embedding obtained via a SentenceTransformer~\cite{reimers2019sentence}:
\[
v_{e_i} = LM(t_{e_i}), \quad v_{r_k} = LM(t_{r_k}),
\]
where $t_{e_i}$ and $t_{r_k}$ denote short textual descriptions extracted by GraphRAG. 
The resulting entity graph serves as the semantic backbone for subsequent reasoning.  

\textbf{Step 2: Query node insertion.}  
For each user query, we introduce a \textbf{Query Node} $q$ with its textual content $t_q$ encoded by the same language model:
\[
v_q = LM(t_q).
\]
We connect $q$ to all entities it references, forming edges $(q, e_j, r_q)$ where $r_q$ is a special ``query-related'' relation with embedding:
\[
v_{r_q} = LM(\text{``This entity is related to this question''}).
\]
This produces a heterogeneous graph $G_q = (V_{G_q}, E_{G_q})$ that includes both query and entity nodes:
\[
V_{G_{q_2}} = \{e_1, e_4, e_5, e_6, q_2\},\]
\[
E_{G_{q_2}} = \{(e_1, e_4, r_3), (q_2, e_1, r_q), (q_2, e_4, r_q), \dots \}.
\]
This design provides a structured and interpretable foundation for multi-hop reasoning across turns.

\textbf{Step 3: Cross-turn graph merging.}  
To detect adversarial intent that gradually emerges across dialogue turns, 
we merge the current query graph $G_{q_t}$ with the global graph $G_{\text{global}}^{(t-1)}$ accumulated from previous turns. 
The merging operation performs a \textit{node-wise union} of entities and relations, 
reusing nodes that share identical entity identifiers or high semantic similarity (cosine similarity above a threshold $\tau$). 
Formally:
\[
G_{\text{global}}^{(t)} = \text{Graph\_Merge}(G_{\text{global}}^{(t-1)}, G_{q_t}),
\]
where duplicate entity nodes are merged, and their associated edges and relation weights are aggregated. 
This process preserves coreference consistency across turns and connects semantically related entities between user queries.  

The resulting global heterogeneous graph $G_{\text{global}}^{(t)}$ captures 
cross-turn dependencies and the evolving conversational context. 
It serves as the final input to the GNN, which classifies whether the current Query Node $q_t$ represents a harmful intent.

\subsection{Attention-aware Augmentation}

Jailbreak attacks exhibit high semantic and structural variability, making it difficult for the GNN to generalize to unseen queries. To address this limitation, we propose an attention-aware augmentation mechanism that improves the robustness and transferability of the model.

Our augmentation method leverages a curated dataset of labeled, single-turn queries. For each incoming multi-turn query sequence, we retrieve the most semantically similar entry from this dataset and insert it into the query graph as a labeled node. This auxiliary supervision guides the GNN in making more informed classification decisions.

\textbf{Step 1: Simple query database construction.}  
We construct a reference dataset $\mathcal{S}$ containing labeled simple queries, each comprising a short, single-turn sentence annotated as \texttt{harmful} or \texttt{benign}. The samples are collected from existing benchmarks such as HarmBench, JailbreakBench, and a GPT-4o-generated prompt set. This database serves as a repository of attack forms.


\textbf{Step 2: Similarity-based retrieval.}  
Given a multi-turn query sequence $Q = \{q_1, q_2, \dots, q_T\}$, we first encode each query $q_i$ into a vector $v_i$ using a sentence encoder. Then, we apply a sentence-level attention mechanism to compute an aggregated representation $v_{\text{agg}}$:
\[
v_{\text{agg}} = \sum_{i=1}^{T} \alpha_i v_i
\]
where the attention weight $\alpha_i$ reflects the relative importance of the $i^{th}$ turn.

Using this representation, we retrieve the most semantically similar query $s^* \in \mathcal{S}$ from the simple query dataset based on cosine similarity:
\[
s^* = \arg\max_{s_i \in \mathcal{S}} \frac{v_{\text{agg}} \cdot s_i}{\|v_{\text{agg}}\| \cdot \|s_i\|}
\]

\textbf{Step 3: Labeled node insertion.}  
We embed $s^*$ as a labeled node $l$ and insert it into the query graph. This node is prefixed with a structured label (e.g., \texttt{"It is harmful: ..."}) and connected to the current Query Node with an edge whose weight corresponds to their similarity score. By introducing $l$, we inject prior knowledge into the GNN’s reasoning space, improving its ability to classify novel or ambiguous queries.

This augmentation mechanism enhances G-Guard's capability to generalize across diverse jailbreak formulations while preserving computational efficiency.

\subsection{Subgraph Selection}
\begin{figure}
  \centering
  \includegraphics[scale=0.58]{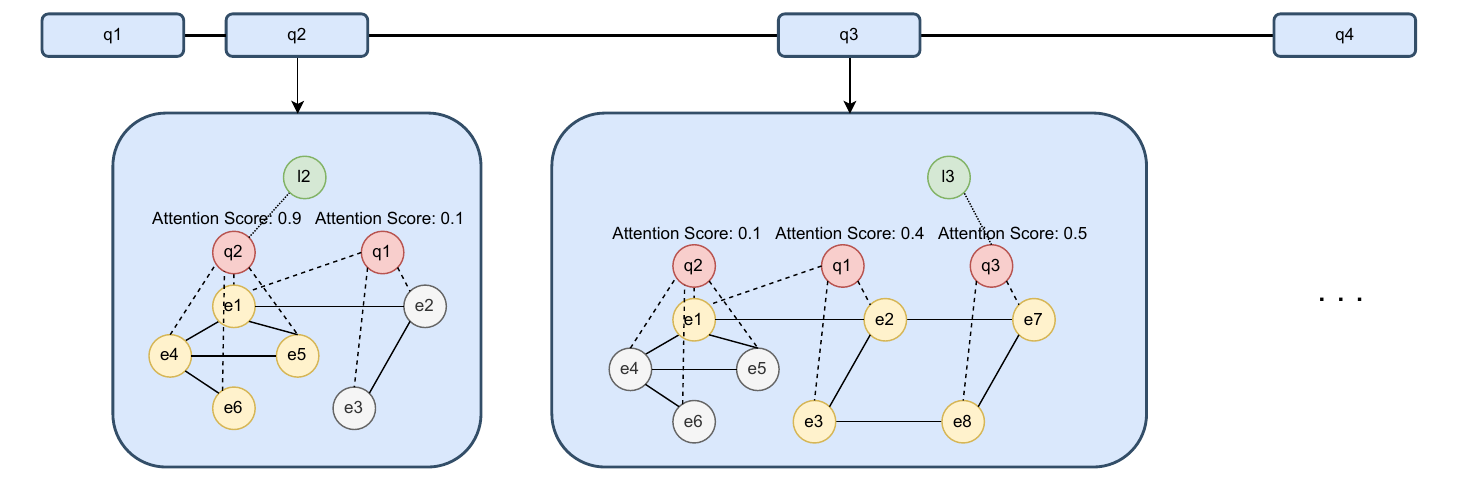}
  \caption{Subgraph Selection. For each incoming query, G-Guard selects a local subgraph from the global graph based on attention scores.}
  \label{selection}
\vspace{-0.7cm}
\end{figure}

As the conversation progresses, the global heterogeneous graph continuously accumulates new nodes from successive queries, 
which may lead to significant computational and memory overhead. 
To ensure scalability, we introduce an \textbf{attention-guided subgraph selection} mechanism 
that dynamically prunes the graph based on query relevance.

When the total number of nodes exceeds a predefined threshold $N_{\text{max}}$, 
we construct a reduced subgraph that preserves only the most informative entities. 
Formally, $N_{\text{max}}$ defines the maximum number of nodes maintained for inference at any given step.

\textbf{Selection policy.}  
We first rank all Query Nodes $\{q_i\}$ by their attention scores $\{\alpha_i\}$, which measure contextual importance. 
We then select the top-$t$ queries with the highest $\alpha_i$ values and retain only their directly connected Entity Nodes. 
If the resulting subgraph still exceeds $N_{\text{max}}$, 
we iteratively reduce $t$ until the node count constraint is satisfied:
\[
G_{\text{sub}} = \{ v \in V_{\text{global}}^{(t)} \mid v \in \mathcal{N}(q_i),\ q_i \in \text{Top-}t(\alpha) \},
\]
where $\mathcal{N}(q_i)$ denotes the set of entity neighbors connected to query $q_i$.

As illustrated in Figure~\ref{selection}, 
when query $q_2$ comes in, the graph exceeds the node limit. while query $q_2$ has a high attention score (e.g., $\alpha_2 = 0.9$) , it
retains its connected entities ($e_1, e_4, e_5, e_6$). 
Later, if new queries arrive and the total node count surpasses $N_{\text{max}}$, we compute the attention score again and 
queries with lower attention (e.g., $\alpha_2 = 0.1$) and their exclusive neighbors 
are removed, thereby reducing graph size.

This hierarchical attention-based pruning policy ensures that the retained subgraph 
emphasizes the most relevant query--entity relationships, 
achieving a favorable trade-off between accuracy, latency, and memory efficiency. 

\section{Experiments}

\subsection{Evaluation Setup}

\textbf{Hardware Configuration.}  
All experiments are conducted on a workstation equipped with an NVIDIA A100 GPU and a 10-core Intel Xeon Cascade Lake CPU. 
The software environment is built on \texttt{Ubuntu 22.04} with \texttt{Python 3.12}, 
\texttt{PyTorch 2.3.0}, and \texttt{CUDA 12.1}. 

\textbf{Attack Methods.}
We evaluate our defense under two state-of-the-art multi-turn jailbreak attack generation techniques: \textbf{Crescendo}~\cite{russinovich2024great} and \textbf{ActorAttack}~\cite{ren2024derail}. Both methods gradually transform benign prompts into adversarial ones through multi-turn interactions, effectively simulating real-world jailbreak scenarios where attackers exploit dialogue history to circumvent LLM safety mechanisms.

\noindent\textbf{Defense Methods.} 
We compare \textbf{G-Guard} with a comprehensive set of existing jailbreak defense mechanisms at the conversation level. 
For each dialogue, the goal is to determine whether any prompts within the conversation are harmful. 

For single-turn baselines, we concatenate all turns into a single input prompt and feed it to the corresponding LLM for evaluation. \textbf{ChatGPT-4o}: OpenAI’s built-in moderation system.
\textbf{Llama Guard 3}~\cite{inan2023llama}: Llama-Guard-3-8B, a rule-based moderation model for detecting harmful content.  
\textbf{Llama Guard 3 FT}: A finetuned version of Llama Guard 3 trained on our multi-turn jailbreak dataset to enhance robustness against contextual attacks.  
\textbf{WildGuard}~\cite{han2024wildguard}: A moderation tool designed to detect harmful prompts and risky user intent; for consistency, we evaluate only its user-prompt moderation component.  
\textbf{ThinkGuard}~\cite{wen2025thinkguard}: A critique-augmented moderation model; we use its prompt-level classification outputs for fair comparison.

For multi-turn conversational defenses, we use Llama-2 as the base model.  
\textbf{Self-Reminder}~\cite{xie2023defending}: A self-regulation mechanism that periodically reminds the model of safety policies during multi-turn interactions, reducing the risk of gradual jailbreaks.  
\textbf{SafeDecoding}~\cite{xu2024safedecoding}: A decoding-level defense that applies safety constraints during generation to prevent harmful continuations at the token level.  
\textbf{G-Guard w/o Aug.}: A variant of our framework without attention-aware retrieval augmentation, used to isolate the contribution of contextual augmentation.  
\textbf{G-Guard}: Our full model integrating attention-aware graph reasoning and retrieval-based augmentation.

\subsection{Dataset}

We evaluate model performance on both public and proprietary datasets: \textbf{HarmBench}~\cite{mazeika2024harmbench} and \textbf{JailbreakBench}~\cite{chao2024jailbreakbench}: These datasets include 400 single-turn prompts (100 benign, 300 harmful). \textbf{ChatGPT-4o-generated dataset:} We create 600 single-turn prompts (300 benign, 300 harmful) across illegal domains. For each single-turn attack, we generate 5 multi-turn jailbreak attacks using ActorAttack and 1 multi-turn jailbreak attack using Crescendo.

\begin{table*}[t]
    \scriptsize
    \centering
        \begin{tabular}{lc|c|c|c|c|c|c}
\toprule
\multirow{2}{*}{\textbf{Method}} &
& \multicolumn{2}{c|}{\textbf{Harmbench}} 
& \multicolumn{2}{c|}{\textbf{JailbreakBench}} 
& \multicolumn{2}{c}{\textbf{ChatGPT-generated}} \\ &
& ActorAttack & Crescendo & ActorAttack & Crescendo & ActorAttack & Crescendo \\

            \midrule
ChatGPT-4o & Accuracy & 0.07 & 0.59 & 0.49 & 0.27 & 0.38 & 0.60 \\
& F1-score & 0.13 & 0.33 & 0.01 & 0.43 & 0.11 & 0.30 \\
            \midrule
Llama Guard 3 & Accuracy & 0.00 & 0.05 & 0.46 & 0.11 & 0.46 & 0.02 \\
& F1-score & 0.00 & 0.09 & 0.00 & 0.19 & 0.00 & 0.03 \\
Llama Guard 3 FT & Accuracy & 0.97 & 0.50 & 0.61 & 1.00 & 0.78 & 0.53 \\
& F1-score & 0.99 & 0.66 & \textbf{0.73} & 1.00 & 0.82 & 0.68 \\
            \midrule
WildGuard & Accuracy & 0.02 & 0.59 & 0.50 & 0.22 & 0.36 & 0.58 \\
& F1-score & 0.04 & 0.33 & 0.00 & 0.36 & 0.02 & 0.24 \\
ThinkGuard & Accuracy & 0.03 & 0.56 & 0.46 & 0.15 & 0.48 & 0.57 \\
& F1-score & 0.06 & 0.27 & 0.03 & 0.26 & 0.09 & 0.28 \\
            \midrule
SafeDecoding & Accuracy & - & 0.63 & - & 0.37 & - & 0.71 \\
& F1-score & - & 0.48 & - & 0.54 & - & 0.59 \\
Self-Reminder & Accuracy & - & 0.70 & - & 0.43 & - & 0.71 \\
& F1-score & - & 0.61 & - & 0.60 & - & 0.59 \\
            \midrule
G-Guard w/o Aug. & Accuracy & 0.71 & 0.73 & 0.53 & 0.77 & 0.84 & 0.84 \\
& F1-score & 0.71 & 0.84 & 0.54 & 0.87 & 0.86 & 0.86 \\
G-Guard & Accuracy & 1.00 & 0.77 & 0.58 & 1.00 & 0.87 & 0.92 \\
& F1-score & \textbf{1.00} & \textbf{0.87} & 0.71 & \textbf{1.00} & \textbf{0.89} & \textbf{0.95} \\

            \bottomrule
        \end{tabular}
    \caption{Overall Performance. Overall performance of G-Guard and baseline methods against ActorAttack and Crescendo on three datasets. Bold values denote the best performance. G-Guard consistently outperforms almost all baselines across accuracy and F1-score.}
    \label{tab:ASR}
\end{table*}

\subsection{Performance}
\begin{figure*}[]
	\centering
	\begin{subfigure}{0.48\linewidth}
		\centering
		\includegraphics[width=0.9\linewidth]{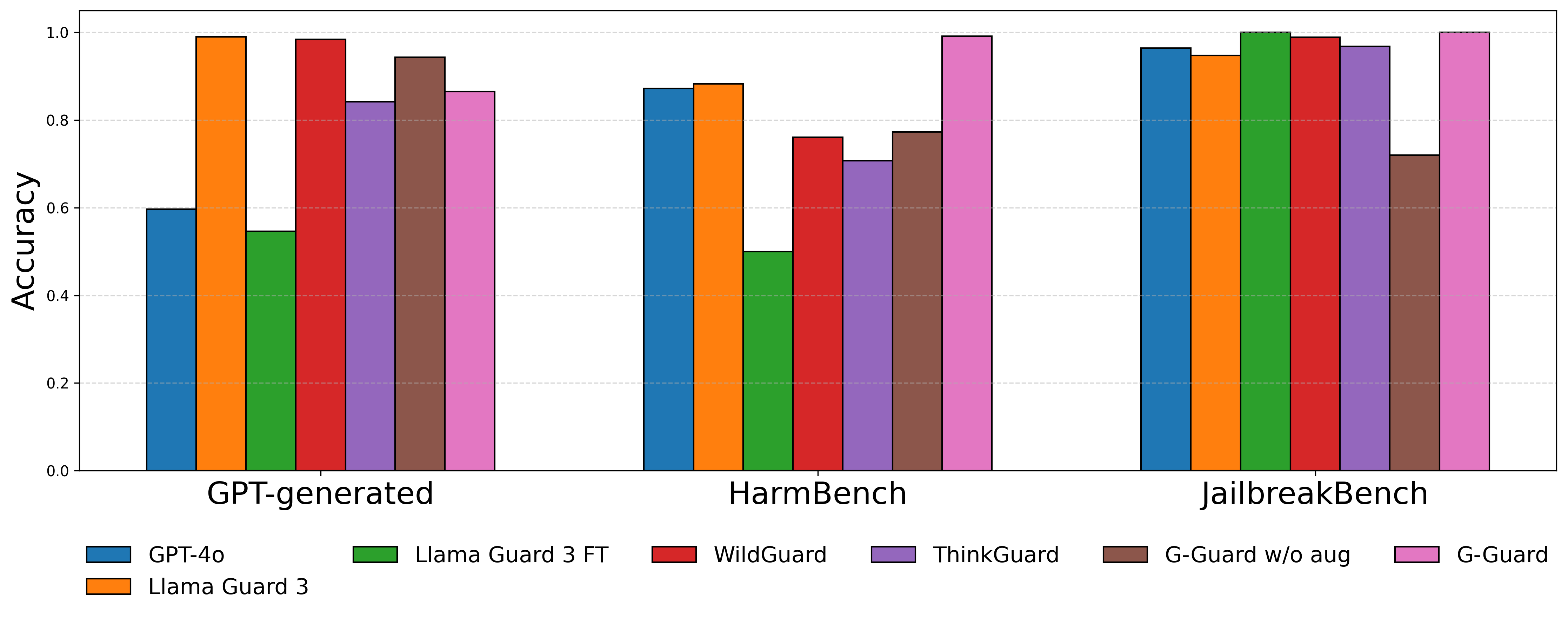}
		\caption{Accuracy in single-turn attacks.}
        \label{single}
	\end{subfigure}
	\centering
	\begin{subfigure}{0.48\linewidth}
		\centering
		\includegraphics[width=0.9\linewidth]{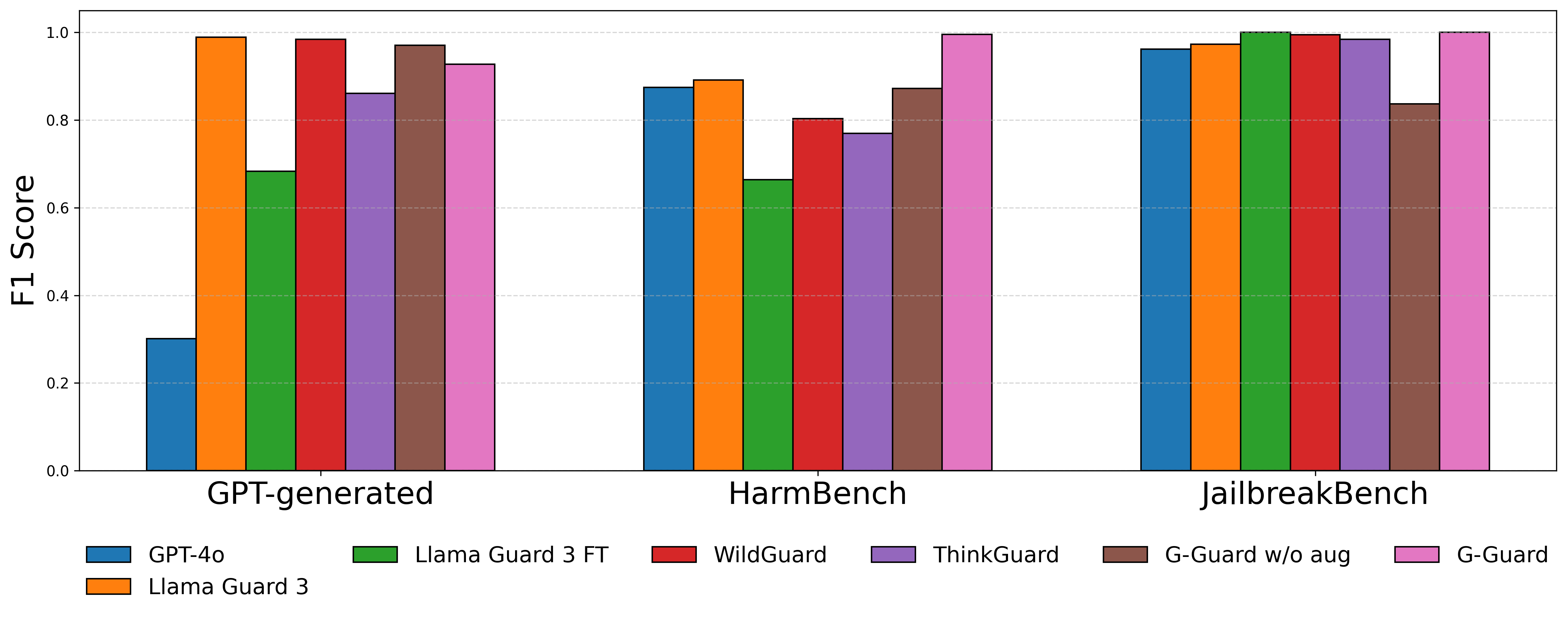}
		\caption{F1-score in single-turn attacks.}
        \label{single}
	\end{subfigure}
    \caption{Single-turn Attack Performance. Accuracy and F1-score of G-Guard and baselines under single-turn jailbreak attacks. G-Guard performs competitively across all datasets despite being designed for multi-turn scenarios.}
    \vspace{-0.5cm}
    \label{fig:single_turn}
\end{figure*}
We first compare G-Guard against a comprehensive set of baseline methods and then analyze the underlying factors contributing to its performance gains. We adopt standard classification metrics: \textit{Accuracy} and \textit{F1-score}, while detailed \textit{Precision} and \textit{Recall} results are provided in the Appendix.

\noindent\textbf{Overall Performance.}
As shown in Table~\ref{tab:ASR}, \textbf{G-Guard consistently outperforms almost all baseline models across all datasets and attack types}. 

On the \textbf{HarmBench} dataset, G-Guard achieves \textbf{perfect accuracy and F1-score (1.00)} under the ActorAttack setting and maintains strong performance (\textbf{0.87} F1-score) under Crescendo. 
This demonstrates its ability to detect gradually emerging harmful intent with no false positives.

On \textbf{JailbreakBench}, G-Guard attains an F1-score of \textbf{0.71} under ActorAttack and \textbf{1.00} under Crescendo, performing on par with Llama Guard 3 FT and surpassing all other defenses. 
This demonstrates G-Guard’s superior ability to capture cross-turn contextual dependencies that elude single-turn or generation-level models.

On the more diverse \textbf{ChatGPT-generated} dataset, G-Guard sustains strong generalization with F1-scores of \textbf{0.89} (ActorAttack) and \textbf{0.95} (Crescendo), outperforming all baselines by a clear margin. 
Even compared to SafeDecoding (\textbf{0.59}) and Self-Reminder (\textbf{0.59}), G-Guard demonstrates superior adaptability.
These results confirm that G-Guard shows the most consistent and robust defense across realistic multi-turn jailbreak scenarios.

\noindent\textbf{Baseline Comparison.}
Llama Guard 3 struggles to detect multi-turn jailbreaks. Llama Guard 3 (Finetuned) demonstrates improved performance, achieving F1 scores of 0.66, 1.00, and 0.68 across three datasets under Crescendo. However, it still falls short compared to G-Guard.

WildGuard exhibits extremely low recall—dropping to as little as 0.02—which results in poor F1-scores. This pattern indicates a highly conservative classification strategy that overlooks a significant number of harmful prompts. ThinkGuard shows comparable behavior, with marginally higher recall, yet continues to underperform in F1-score across all datasets.

ChatGPT-4o’s built-in moderation shows moderate effectiveness, with F1-scores ranging from 0.13 to 0.43. Although it occasionally identifies specific patterns, its inconsistent recall across datasets highlights limitations in managing diverse multi-turn prompting strategies.

\noindent\textbf{Effect of Augmentation.}
G-Guard w/o augmentation achieves solid performance, particularly on HarmBench (F1 = 0.84) and ChatGPT-generated (F1 = 0.86) under Crescendo. However, the full G-Guard (with attention-aware augmentation) improves across all datasets and attack types—raising F1 on Harmbench Crescendo from 0.84 to 0.87 and on ActorAttack from 0.71 to 1.00. This indicates that the retrieved single-turn support examples enhance GNN’s ability to generalize across varied and complex attack forms.

\subsection{Detail Performance}

\begin{figure}[ht]
	\centering
	\begin{subfigure}{0.45\linewidth}
		\centering
		\includegraphics[width=0.9\linewidth]{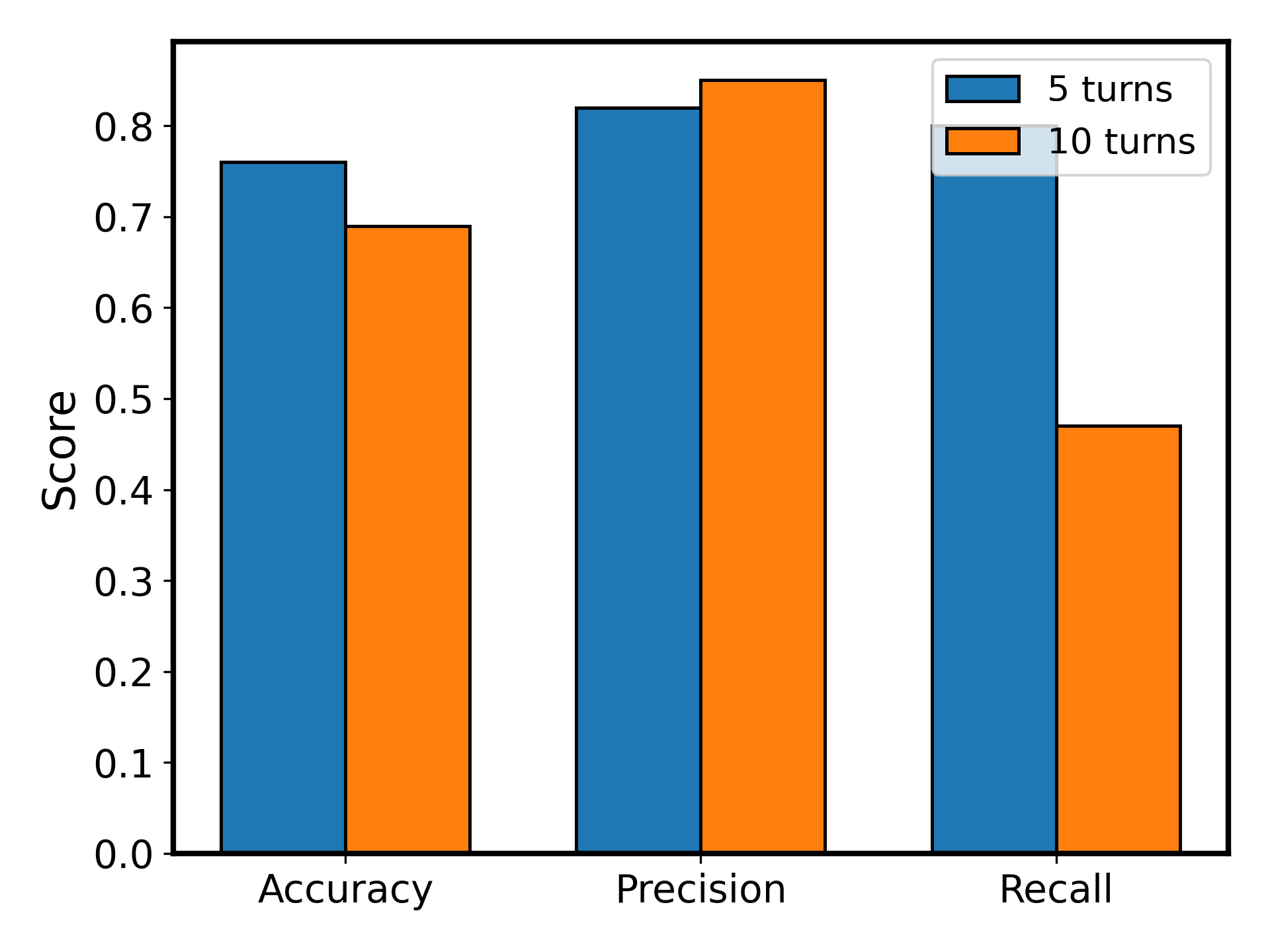}
        \vspace{-0.3cm}
		\caption{Longer turn attacks.}
        \label{long}
	\end{subfigure}
	\centering
	\begin{subfigure}{0.45\linewidth}
		\centering
		\includegraphics[width=0.9\linewidth]{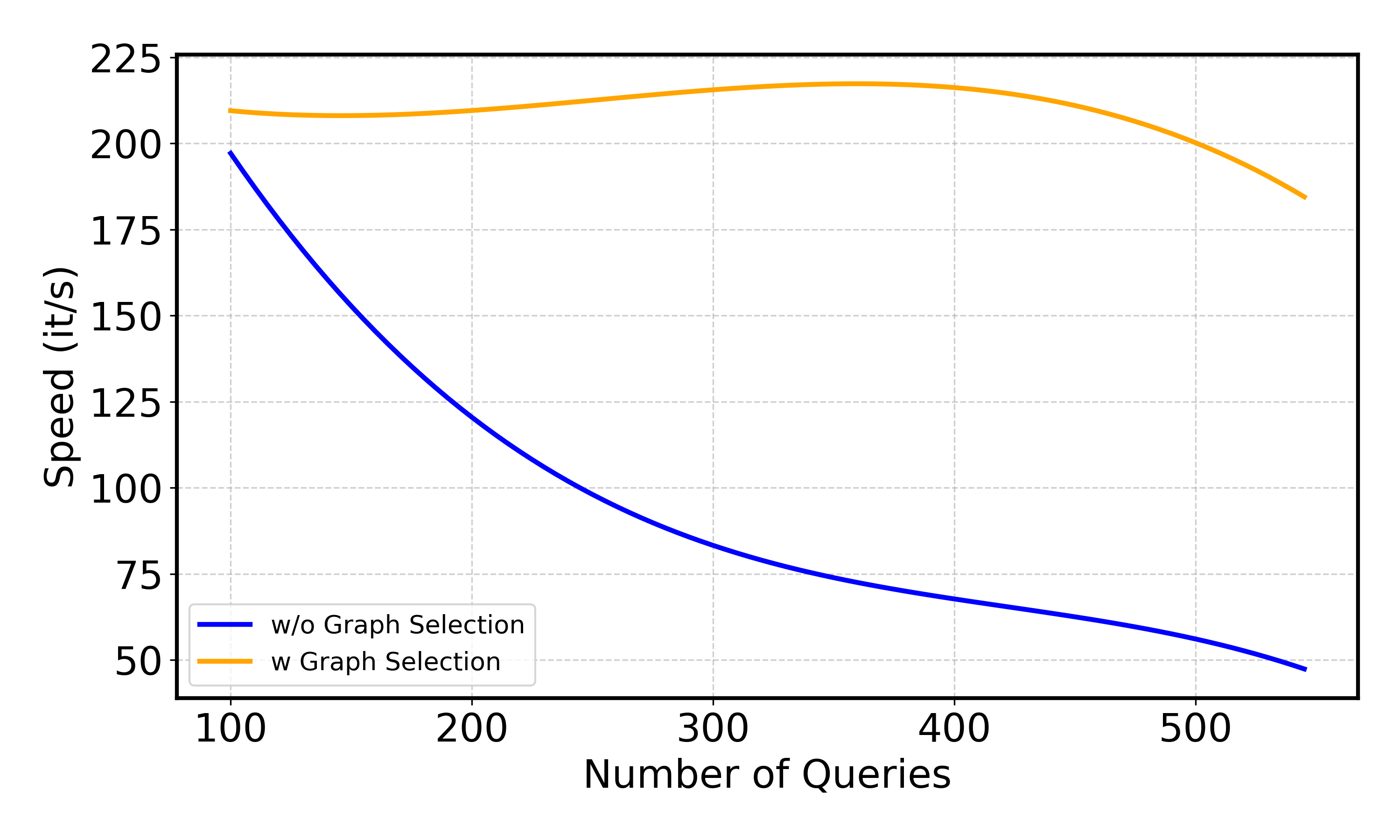}
        \vspace{-0.3cm}
		\caption{More multi-turn attacks.}
        \label{speed}
	\end{subfigure}
    \caption{Detail Performance. G-Guard’s performance under different multi-turn attack settings: (a) longer single multi-turn attacks; (b) multiple simultaneous multi-turn attacks. 
    }
    \vspace{-0.5cm}
\end{figure}
We evaluate G-Guard’s robustness in longer, more realistic conversations under three challenging scenarios: (1) single-turn jailbreak attack, (2) single multi-turn attack distributed across more queries, and (3) multiple distinct multi-turn attacks occurring within the same conversation.

\noindent\textbf{Single-turn Attack.}
Although G-Guard is specifically designed to defend against multi-turn jailbreak attacks, it also generalizes well to single-turn scenarios. As shown in Figure~\ref{fig:single_turn}, G-Guard achieves competitive or superior performance across all three datasets (GPT, HarmBench, and JailbreakBench) in both accuracy and F1-score.

On the GPT dataset, most baseline methods perform well, with G-Guard still maintaining comparable accuracy and F1-score. This confirms that our graph-based reasoning and augmentation mechanisms do not hurt generalization.

For HarmBench dataset, G-Guard shows clear advantages, outperforming Llama Guard 3 (finetuned) and ThinkGuard in both metrics. In particular, WildGuard struggles with low accuracy due to its overly conservative classification, whereas G-Guard balances both precision and recall.

On the JailbreakBench dataset, G-Guard again ranks among others. Notably, the augmented version (G-Guard) outperforms the variant without augmentation (G-Guard w/o aug), suggesting that the retrieved labeled queries also help in capturing implicit harmful intent in single-turn inputs.

This result demonstrates that although G-Guard is built for multi-turn defense, its architecture and augmentation pipeline remain effective in traditional single-turn threat models, eliminating the need for separate classifiers.

\noindent\textbf{Effect of Graph Construction.}
We show the effect of our graph construction method in the JailbreakBench dataset with Actorattack. As shown in Figure \ref{long}, with graph construction, G-Guard continues to perform well even as the conversation length increases from 5 to 10 turns. Despite a slight drop in recall—attributable to the growing complexity and fragmentation of adversarial intent in extended interactions—G-Guard maintains high accuracy, precision, and F1 scores. This suggests it consistently detects harmful queries, while effectively minimizing false positives. This shows that our method maintains strong robustness and effectiveness in longer multi-turn settings.

\noindent\textbf{Effect of Graph Selecting.}
We show the effect of our graph selecting method in the last scenarios. Figure \ref{speed} compares the inference speed of our method with and without the proposed graph selection mechanism as the number of queries increases. Without graph selection (blue line), speed drops significantly as the graph grows larger. In contrast, with graph selection (orange line), the speed remains stable and consistently high, even as the number of queries exceeds 500. This demonstrates that our graph selection module effectively controls computational overhead and ensures scalability. Importantly, this optimization comes with little drop in accuracy, confirming that the method preserves essential contextual information.


\section{Conclusion}
In this work, we present G-Guard, a novel defense framework designed to detect and mitigate multi-turn jailbreak attacks on LLMs. Unlike traditional methods that primarily operate on single-turn inputs, G-Guard leverages a graph-based representation to capture the evolving semantics and contextual dependencies across multi-turn conversations. By constructing entity graphs and integrating an attention-aware augmentation mechanism, G-Guard effectively models inter-query relationships and enhances generalization to diverse attack patterns. Our evaluation across multiple benchmark datasets shows that G-Guard consistently outperforms state-of-the-art baselines in terms of accuracy, precision, recall and F1-score.

\bibliography{colm2025_conference}
\bibliographystyle{colm2025_conference}

\appendix
\section{Appendix}
\subsection{Limitations}
While G-Guard demonstrates strong performance in detecting multi-turn jailbreak attacks, it is not without limitations. Below, we discuss two key areas where our method faces challenges: scalability and generalization.
\subsubsection{Scalability}
Although G-Guard uses subgraph selection to maintain a fixed GNN input size, scalability challenges still arise when conversations become extremely long or complex. In such cases, the model must aggressively exclude less relevant nodes to fit within the input size constraint. This may lead to the loss of potentially important contextual information, especially when harmful intent is distributed across many turns. As a result, the model’s ability to capture long-range dependencies may degrade, potentially impacting detection accuracy in very lengthy conversations. While the fixed input size ensures efficiency, it also introduces a trade-off between context preservation and computational feasibility, which remains a limitation in scaling to real-world dialogue systems with highly extended interactions.
\subsubsection{Generalization}
Another limitation lies in the method’s ability to generalize to novel or evolving attack strategies. While G-Guard performs well on known datasets and common multi-turn attack patterns, new jailbreak techniques continue to emerge rapidly—often exploiting unforeseen vulnerabilities, linguistic variations, or prompt injection tricks. Since our model relies on learned patterns and a retrieval-based augmentation approach, its effectiveness against unseen attack types is uncertain. Future work is needed to improve the model's adaptability to evolving adversarial behaviors and reduce reliance on static training data or retrieval corpora.

\subsection{Evaluation Detail}

\begin{table*}[h]
    \scriptsize
    \centering
        \begin{tabular}{lc|c|c|c|c|c|c}
\toprule
\multirow{2}{*}{\textbf{Method}} &
& \multicolumn{2}{c|}{\textbf{Harmbench}} 
& \multicolumn{2}{c|}{\textbf{JailbreakBench}} 
& \multicolumn{2}{c}{\textbf{ChatGPT-generated}} \\ &
& ActorAttack & Crescendo & ActorAttack & Crescendo & ActorAttack & Crescendo \\

            \midrule
ChatGPT-4o & Accuracy & 0.07 & 0.59 & 0.49 & 0.27 & 0.38 & 0.60 \\
& Precision & 1.00 & 0.86 & 0.33 & 1.00 & 0.90 & 1.00 \\
& Recall & 0.07 & 0.20 & 0.01 & 0.27 & 0.06 & 0.18 \\
& F1-score & 0.13 & 0.33 & 0.01 & 0.43 & 0.11 & 0.30 \\
            \midrule
Llama Guard 3 & Accuracy & 0.00 & 0.05 & 0.46 & 0.11 & 0.46 & 0.02 \\
& Precision & 0.00 & 1.00 & 0.00 & 1.00 & 0.00 & 1.00 \\
& Recall & 0.00 & 0.05 & 0.00 & 0.11 & 0.00 & 0.02 \\
& F1-score & 0.00 & 0.09 & 0.00 & 0.19 & 0.00 & 0.03 \\
Llama Guard 3 FT & Accuracy & 0.97 & 0.50 & 0.61 & 1.00 & 0.78 & 0.53 \\
& Precision & 1.00 & 0.50 & 0.59 & 1.00 & 0.71 & 0.51 \\
& Recall & 0.97 & 1.00 & 0.97 & 1.00 & 0.98 & 1.00 \\
& F1-score & 0.99 & 0.66 & \textbf{0.73} & 1.00 & 0.82 & 0.68 \\
            \midrule
WildGuard & Accuracy & 0.02 & 0.59 & 0.50 & 0.22 & 0.36 & 0.58 \\
& Precision & 1.00 & 0.86 & 0.00 & 1.00 & 1.00 & 1.00 \\
& Recall & 0.02 & 0.20 & 0.00 & 0.22 & 0.01 & 0.14 \\
& F1-score & 0.04 & 0.33 & 0.00 & 0.36 & 0.02 & 0.24 \\
ThinkGuard & Accuracy & 0.03 & 0.56 & 0.46 & 0.15 & 0.48 & 0.57 \\
& Precision & 1.00 & 0.79 & 1.00 & 1.00 & 0.75 & 0.80 \\
& Recall & 0.03 & 0.16 & 0.02 & 0.15 & 0.05 & 0.17 \\
& F1-score & 0.06 & 0.27 & 0.03 & 0.26 & 0.09 & 0.28 \\
            \midrule
SafeDecoding & Accuracy & - & 0.63 & - & 0.37 & - & 0.71 \\
& Precision & - & 0.78 & - & 1.00 & - & 0.98 \\
& Recall & - & 0.34 & - & 0.37 & - & 0.43 \\
& F1-score & - & 0.48 & - & 0.54 & - & 0.59 \\
Self-Reminder & Accuracy & - & 0.70 & - & 0.43 & - & 0.71 \\
& Precision & - & 0.85 & - & 1.00 & - & 0.98 \\
& Recall & - & 0.47 & - & 0.43 & - & 0.43 \\
& F1-score & - & 0.61 & - & 0.60 & - & 0.59 \\
            \midrule
G-Guard w/o Aug. & Accuracy & 0.71 & 0.73 & 0.53 & 0.77 & 0.84 & 0.84 \\
& Precision & 1.00 & 0.78 & 0.56 & 1.00 & 0.82 & 0.82 \\
& Recall & 0.71 & 0.90 & 0.53 & 0.77 & 0.91 & 0.91 \\
& F1-score & 0.71 & 0.84 & 0.54 & 0.87 & 0.86 & 0.86 \\
G-Guard & Accuracy & 1.00 & 0.77 & 0.58 & 1.00 & 0.87 & 0.92 \\
& Precision & 1.00 & 0.77 & 0.56 & 1.00 & 0.81 & 0.91 \\
& Recall & 1.00 & 1.00 & 1.00 & 1.00 & 1.00 & 1.00 \\
& F1-score & \textbf{1.00} & \textbf{0.87} & 0.71 & \textbf{1.00} & \textbf{0.89} & \textbf{0.95} \\

            \bottomrule
        \end{tabular}
    \caption{Overall Performance. Overall performance of G-Guard and baseline methods against ActorAttack and Crescendo on three datasets. Bold values denote the best performance.}
    \label{tab:ASR2}
\end{table*}
\end{document}